# Image Optimization and Prediction

**Ms. Sweta V. Jain**
M.Tech Final Year, CSE, Hingana Road, Nagpur.
.                                    swetavjain@rediffmail.com

**Urmila Shrawankar**
**Assistant Professor, Computer Science and Engineering Department**
G.H.R.C.E., Hingana Road, Nagpur
urmila@ieee.org

**Abstract**

Image Processing, Optimization and Prediction of an Image play a key role in Computer Science. Image processing provides a way to analyze and identify an image .Many areas like medical image processing, Satellite images, natural images and  artificial images requires  lots of analysis and research on optimization. In Image Optimization and Prediction we are combining the features of Query Optimization , Image Processing  and Prediction .

Image optimization is used in Pattern analysis, object recognition ,in   medical Image processing to predict the type of diseases ,in satellite images for predicting weather forecast ,availability of water or mineral etc. Image Processing , Optimization and  analysis  is  a wide open area for research .Lots of research has been conducted in the area of Image analysis and many techniques are available for image analysis but , a  single  technique is not yet identified  for  image analysis and prediction .our research is focused on identifying a global technique for image analysis and Prediction.

## 1 Introduction

Typical computer vision applications usually require an image segmentation-preprocessing algorithm as a first procedure. At the output of this stage, each object of the image, represented by a set of pixels, is isolated from the rest of the scene. The purpose of this step is that objects and background are separated into non overlapping sets. Usually, this segmentation process is based on the image gray-level histogram. In that case, the aim is to find a critical value or threshold. Through this threshold, applied to the whole image, pixels whose gray levels exceed this critical value are assigned to one set and the rest to the other. For a well-defined image, its histogram has a deep valley between two peaks. Around these peaks the object and background gray levels are concentrated. Thus, to segment the image using some histogram thresholding technique, the optimum threshold value must be located in the valley region. A myriad of algorithms for histogram thresholding can be found in the literature.

During last decades, growing attention has been put on data clustering as robust technique in data analysis. Clustering or data grouping describes important technique of unsupervised classification that arranges pattern data (most often vectors in multidimensional space) in the clusters (or groups). Patterns or vectors in the same cluster are similar according to predefined criteria, in contrast to distinct patterns from different clusters .Possible areas of application of clustering algorithms include data mining, statistical data analysis, compression, vector quantization and pattern recognition [12]. Image analysis is the area where grouping data into meaningful regions (image segmentation) presents the first step into more detailed routines and procedures in computer vision and image understanding. Clustering problem understood as grouping input data by means of minimizing certain criteria presents NP-hard combinatorial optimization task. Genetic algorithms are classified as population based optimization techniques that make extensive use of the mechanisms met in evolution and natural genetics. For this reason combining clustering techniques with genetic algorithms robustness in optimization should yield high quality performance and results. The remainder of this paper is organized as follows Section 2 provides related work .Section 3 describes Methodology. Section 4 describes Result & Disscussion .Section 5 Concludes the paper and point Future Scope.

**2 Related Work**

In Image segmentation and processing various techniques are available like Thresholding, Histogram, Region growing, Clustering, Split and Merge, Watershed algorithm, wavelet based segmentation, Interactive Image segmentation using adaptive weighted distance, Image segmentation with scalable spatial information, Neural network based segmentation [1]- [19] [24],[25] .It has been observed that some techniques are suitable for grey scale images while others for color images, high intensity image, satellite images and so on. There is no such global technique that is suitable for variety of application domains .

Primary image segmentation is defined as an optimal segmentation obtained in a pure bottom-up fashion that provides the information necessary to initialize and constrain high-level segmentation methods. Although the details of primary segmentation methods will depend on the application domain, we require that they do not depend on a priori knowledge about the objects present in a particular scene or image specific parameter adjustments. These claims become realistic because we do not seek for a perfect segmentation result but rather for the best possible support for more intelligent methods to be applied afterwards. Unfortunately up to now there is no theory which defines the quality of segmentation. Therefore we have to rely on some heuristic constraints which the primary segmentation should meet

- The segmentation should provide regions that are homogenous with respect to one or more properties, i.e. the variation of measurements within the regions should be considerably less than the variation at borders.
- The position of the borders should coincide with local maxima, ridges and saddle points of the local gradient of the measurements.
- -Areas that perceptually form only one region should not be splitted into several parts. In particular this applies to smooth shading and texture.
- Small details, if clearly distinguished by their shape or contrast, should not be merged with their neighboring regions.

**2.1 Primary Image Segmentation**

As was already mentioned primary segmentation should be region based segmentation. The simplest idea to obtain this would be the complement of a standard edge detection. Unfortunately edge detection algorithms tend to miss important edges if their contrast is poor which results in an under segmentation. Even an extended algorithm that tries to close edge fragments starting from terminating points as proposed will not reliably find all region boundaries.

Here the segmentation process is split into two stages: at first an initial incomplete segmentation forming seed regions is performed. A subsequent region growing stage assigns the yet unlabeled pixels to one of the regions until the segmentation is complete. This leads to a four-step framework for primary segmentation : preprocessing, seed selection, region growing and optional post processing to improve the regions.

Scene analysis is a central task in numerous applications including autonomous robots, intelligent human computer interfaces, and content-based image and video retrieval systems. Its essential goal is to derive a meaningful description of the input. While a hypothetical solution can be constructed by building a look-up table with one entry for each possible input, the complexity of labeling the table entries, storing the table, and finding a match for a given input makes it impossible to implement. Clearly, the complexity is a key requirement for scene analysis and the focus of research is thus on developing efficient and effective models and algorithms. To illustrate the plausible approaches to scene analysis, Fig. 1(a) shows a natural image of a cheetah. One choice is to detect objects by exhaustively searching over all possible variations, including locations, scales, and orientations; this strategy is used by some face detection algorithms .This approach, however, is not effective given the large number of possible objects and their variations in the input. It seems that a more promising approach is to decompose scene analysis into two stages: an initial generic scene segmentation followed by some iterative recognition and model-specific segmentation loop. Note that the initial segmentation stage is critical for the decomposition to be effective; otherwise, it would reduce to exhaustive detection. In this setting, the goal of scene segmentation is not to obtain an ideal segmentation—in fact, it was argued that an ideal segmentation like the one shown in Fig. 1(c) is not feasible without specific models —but to provide candidate regions to initiate recognition, classification, or other higher level processes. This formulation makes clear the goal of generic scene segmentation, as well as its constraints. Clearly, generic scene segmentation should not utilize object-specific models as they are not available. Also, a solution should work on different kinds of images such as texture and non texture images. There are two broad categories of approaches to bottom-up scene segmentation. The first category is to group basic elements that can be computed easily (e.g., edges) to form more meaningful boundaries. However, meaningful elements may not be obtained readily due to complex textures of natural objects. For example, Fig. 1(b) shows a typical edge-detection result for Fig. 1(a) using the Canny edge detector . It is clear that grouping these edges into meaningful boundary segments is a complicated task, if not infeasible. The other category is to identify regions based on statistics of local features, including region growing, split-and merge, and so on. Here, segmentation can be defined as a constrained partition problem, where each partitioned region should be as homogeneous as possible and neighboring ones should be as different as possible. Two fundamental issues can be identified: The first one is the underlying image model that defines region homogeneity and thus specifies what a good segmentation should be, and the second one is to design an algorithm to compute a solution, exactly or approximately. There are numerous algorithms for segmentation. A key difference among them is the underlying segmentation criterion, either explicitly or implicitly defined. Assuming that a representative spectral histogram is available for each region, we derive a segmentation algorithm as follows: 1) estimating a probability model for each region and classifying image windows to obtain an initial segmentation; 2) iteratively updating segmentation and local spectral histograms of pixels along region boundaries based on the derived probability models; and 3) further localizing region boundaries using refined probability models derived based on spatial patterns in segmented regions[2].

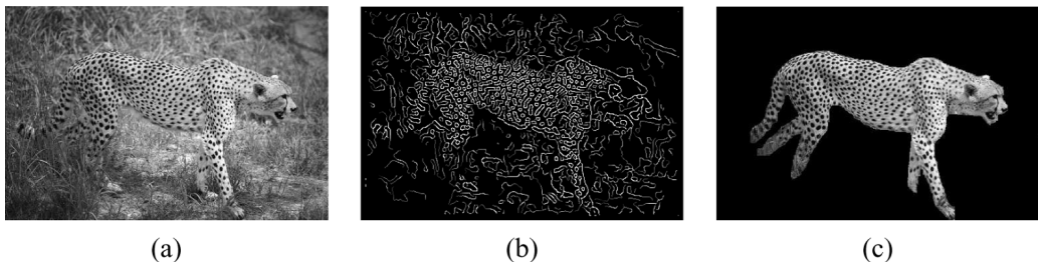

(a)                  (b)                  (c)

Fig. 1. Illustrative example. (a) Natural image of a cheetah. (b) Typical edge detection result. (c) Manually segmented result.

## 3 Methodology

### 3.1 Standard *K*-Means Clustering Algorithms

Clustering techniques usually are divided into two general groups: hierarchical and partitional clustering algorithms. Hierarchical clustering techniques create a cluster tree by means of heuristic splitting or merging procedures. On the other hand, partitional clustering techniques divide the input data into specified in advance number of clusters. The whole process is governed by minimization of certain goal function, e.g. a square error function. "Center-based clustering" refers to the family of algorithms that use a number of "centers" to represent and group input data. General iterative model for partitional center-based clustering algorithms has the following form [12]:
1. Data initialization by assigning some values to the cluster centers.
2. For each data point xi, calculate its membership value m(c j | xi ) to all clusters c j and its weight w(xi ) .
3. For each cluster center cj, recalculate its location taking into account all points xi assigned to this cluster according to the membership and weight values:

$$c_j = \frac{\sum_{i=1}^{n} m(c_j | x_i) w(x_i) x_i}{\sum_{i=1}^{n} m(c_j | x_i) w(x_i)}$$

4. Repeat steps 2 and 3 until some termination criteria are met. Standard *k*-means clustering algorithms require that cluster number *k* should be determined in advance. Additionally, results (segmentations) obtained in the run of the *k*-means algorithm strongly depend on the selection of initial clusters centers.
The proposed Methodology for Image Optimization and Prediction requires following

- Query optimization [20-22]
- Image processing
- Prediction

To design this Software, initially we will keep Images in the Database with their short text description . The Image to be analyzed and Predicted will be input to the system. The Input Image is segmented using Image Segmentation technique .The segmentation of image will help us to analyze the image .Query will be used to retrieve the images which are matching with input image.This Query will generate many results but, we will apply Query optimization technique to find the best promising results. Finally we will relate all the matching parts and we can apply neural network or Fuzzy logic rule base to identify the meaning of an Image . we can use this information for prediction of an image .
This project is inspired from an Original Idea ,till now image analysis , segmentation , query optimization and object recognition has been explored individually .our methodology is to combine image segmentation ,analysis, query optimization together for prediction of an image .

**4 Result and Discussion**

**As** an illustration of the differences between the K-means and edge adaptive K-means, both algorithms were applied to the Lena image.

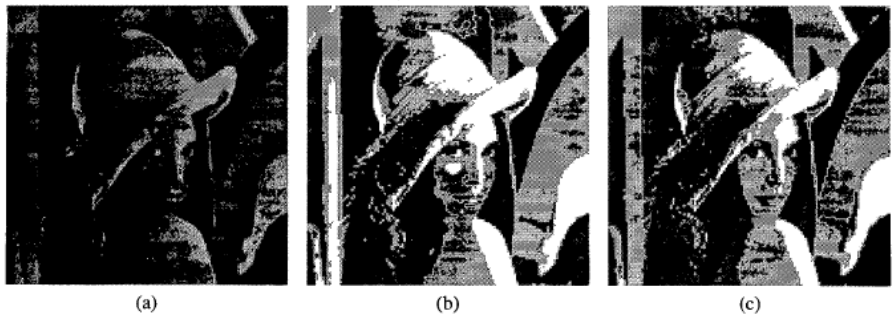

**Fig. 2.** Segmentation of Lena image into four classes: (a) original image, (b) standard K-means segmentation, (c) edge-adaptive segmentation.

Figure 2(a) shows the original image ,Figure 2(b) shows the standard K-means result, and Figure2(c) shows the edge-adaptive K-means result. Four classes were arbitrarily assumed. The differences between the two results are fairly subtle. However, it can be seen that the edge-adaptive segmentation does not possess the additional white region within the vertical bar on the left side of the image, nor does it possess the white patch on the left side of the face. This is a result of the edge adaptive algorithm's tradeoff between reconstructing the intensity values in the orignal image versus reconstructing the boundary details. **A** strong enough edge was not found to cause those pixels to be in a different class. More detail within the hair is also evident in the edge-adaptive result.

**5 Conclusion and Future Scope**

Clustering algorithms are widely used due to unsupervised learning ,further the training data set is not required. The edge-adaptive clustering algorithm can be easily extended to three-dimensional volumetric images, as well as multi-channel images. In the latter case, a suitable gradient operator for vector valued functions needs to be defined.

- Image optimization technique will be used in Pattern analysis, object recognition, in medical Image processing to predict the type of diseases [25].
- In satellite images for predicting weather forecast, availability of water or mineral, agriculture, fisheries, watershed development, water land mapping, monitoring irrigated commands.
- Automatic analysis of remote sensing data from satellites to identify and measure regions of interest. e.g. Petroleum reserves.